\documentclass[runningheads]{llncs}
\usepackage[T1]{fontenc}
\usepackage{amsmath} 
\usepackage{graphicx}
\usepackage{amsmath,amssymb}
\usepackage{booktabs}
\usepackage{algorithm}
\usepackage{algorithmic}
\usepackage[table]{xcolor}
\usepackage{color}
\begin{document}
\title{GSAM: A Generalizable and Safe Robotic Framework for Articulated Object Manipulation}
\titlerunning{Generalizable and Safe Articulated Object Manipulation}
% If the paper title is too long for the running head, you can set
% an abbreviated paper title here
%
\author{Beichen Shao\inst{1}\and
Mengying Xie\inst{1}\and
Heng Su\inst{1}\and
Wanyi Zhang\inst{1}\and
Mingyan Li\inst{1}\and
Yan Ding\inst{2,3,4}\and
Fausto Giunchiglia\inst{5}\and
Chao Chen\inst{1}
\thanks{Corresponding author: \email{cschaochen@cqu.edu.cn} \\Beichen Shao and Mengying Xie contributed equally to this work.}}
\authorrunning{B. Shao et al.}
% First names are abbreviated in the running head.
% If there are more than two authors, 'et al.' is used.
%
\institute{College of Computer Science, Chongqing University, Chongqing, China
\email{cschaochen@cqu.edu.cn}\\
\and
Lumos Robotics, China
\and
Xi'an Jiaotong-Liverpool University, China 
\and
Fudan University, China
\and
Department of Information Engineering and Computer Science, University of Trento, Trento, Italy
% \email{lncs@springer.com}\\
% \url{http://www.springer.com/gp/computer-science/lncs} 
\\
%\email{\{abc,lncs\}@uni-heidelberg.de}
}
\maketitle              % typeset the header of the contribution
\begin{abstract}
Articulated object manipulation is a unique challenge for service robots. Existing methods employ end-to-end policy learning, vision-motion planning, and large-language/visual-language model (LLM/VLM), but often overlook the \textbf{diversity} of articulated objects and the \textbf{complexity} of interactions between end-effector and handle, leading to limited generalization and destructive collisions. To address this, we propose GSAM, a generalizable and safe robotic framework for articulated object manipulation.
Specifically, a vision-based perceiver generates the kinematic parameters. Considering that pre-trained markers in perceiver yield {\em raw} estimations that may deviate from commonsense, we present a fine-tuned VLM-based refiner, using chain-of-thought (COT) commonsense reasoning to refine perception. To prevent destructive collisions, we design an interaction constraint function generator, integrating articulated object, interaction pose, and obstacle avoidance knowledge into a base. LLM then functionalize these constraints and apply them to trajectory and posture planning. A kinematic-aware manipulation planner verifies reachability for trajectory and posture.
Experiments on 50 hinge tasks across 5 object categories and 50 randomly initialized end-effector-handle configurations show that GSAM reduces standard deviation by 3.1\% and improves manipulation success rate by 36.0\% compared to the best baseline, respectively demonstrating the superior object generalization and interaction safety of GSAM in practical scenarios.

\end{abstract}
\section{Introduction}
Service robots are widely applied in offices, households, and other social environments. Articulated object manipulation is a key capability for it to enable fine-grained operations and human-robot collaboration. However, due to diverse articulated objects and complex interaction relationships between end-effector and handle, such manipulation remains prone to failure with inaccurate kinematic parameter estimation and destructive collisions, making the development of a generalizable and safe manipulation strategy highly valuable.

Prior works on articulted object manupulation utilize the end-to-end policy learning, vision-motion planning, and LLM/VLM. Specifically, end-to-end policy learning methods~\cite{chi2024umi,fan2023digital,wu2024gello,levine2018learning} are tailored to a specific type and size of articulated objects. These methods learn manipulation policies by imitating human demonstrations and using reinforcement learning (RL) to establish behavioral incentives. Vision-motion planning approaches~\cite{gupta2025openingarticulatedobjectsreal,karayiannidis2016adaptive} use visual perception to estimate kinematic parameters, followed by kinematic-aware trajectory planning. LLM/VLM-based methods~\cite{xia2024kinematic,huang2024rekep,wei2022chain} identify keypoints from point cloud or visual data and directly generate manipulation trajectories based on instructions. Leveraging the reasoning of LLMs and VLMs, these methods expect to introduce world knowledge to improve adaptability in real-world tasks.
\begin{figure}
    \centering
    \includegraphics[height=0.25\linewidth]{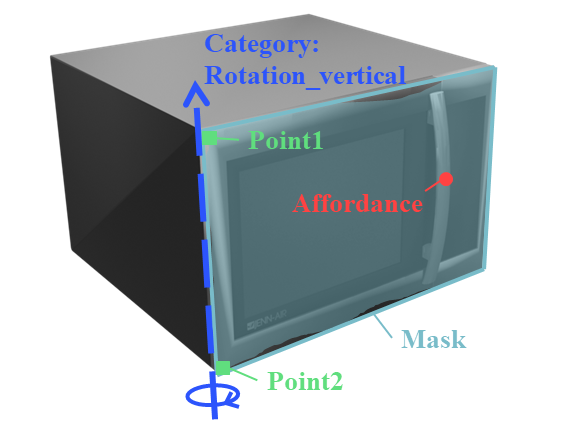}
    \includegraphics[width=0.27\linewidth]{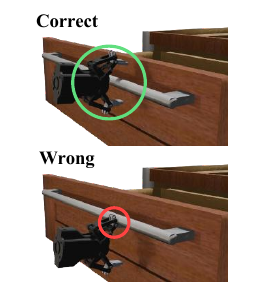}
    \vspace{-6pt}
    \caption{Illustrations of (a) the kinematic parameters of articulated objects and (b) the correct/wrong interaction between end-effectors and handles.}
    \vspace{-15pt}
    \label{fig:kinematic-interaction}
\end{figure}
Despite the effectiveness of the above methods, two critical issues remain to be addressed.

(1) \textit{Generalization challenge due to diverse articulated objects.} As shown in Fig. 1(a), accurate estimation of kinematic parameters, including affordance (interaction position), category (hinge type), masks (object size) and keypoints (hinge location), is vital for manipulation. In practice, these parameters vary significantly across distinct objects. Existing approaches either train object-specific perception policies using imitation or reinforcement learning, or use external vision models without fine-tuning. The former fails to generalize to unseen objects, while the latter often produces the results inconsistent with commonsense. Addressing this generalization challenge remains essential.

(2) \textit{Safety challenge due to complex end-effector-handle interactions.} As illustrated in Fig. 1(b), interaction complexity depends on both the initial pose of the handle and the initial position of the end-effector. End-effectors often fail to grasp handles, especially in narrow spaces, leading to destructive collisions. Ensuring safe interaction requires systematically constructing constraints based on the knowledge of articulated object, interaction pose, and obstacle avoidance. Existing methods focus either on motion feasibility or grasp rule generation, but rarely integrate all knowledge types simultaneously. Thus, safety remains a pressing issue in manipulation planning.

In this study, we propose a \textbf{G}eneralizable and \textbf{S}afe robotic framework for \textbf{A}rticulated object \textbf{M}anipulation (GSAM), which contains four modules. First, GSAM uses Kinematic Parameter Perceiver (KPP) to obtain raw kinematic parameters of articulated objects. Then, in the Kinematic Parameter Refiner (KPR), we employ a fine-tuned VLM to incorporate commonsense knowledge through COT reasoning, thereby enhancing generalization. Next, in interaction Constraint Function Generator (CFG), we construct a knowledge base integrating articulated object, interaction pose and obstacle avoidance information, and employ a fine-tuned LLM to change it into constraint functions for safe interaction. Finally, a Kinematic-aware Manipulation Planner (KMP) ensures reachability in both trajectory and posture, enhancing real executions. 

Our contributions are summarized as follows:
\begin{itemize}
\item We propose GSAM, an innovative robotic framework capable of generalizable object perception and safe end-effector-handle interaction, achieving state-of-the-art performance in articulated object manipulation.
\item We develop the KPR module, which leverages the reasoning capabilities of a fine-tuned VLM to correct commonsense-inconsistent kinematic parameter estimations, thereby enhancing manipulation generalization.
\item We introduce CFG, which presents a structured knowledge base and LLM functionalization to avoid destructive collisions, improving safety in multi-object scenarios.
% \item We conduct extensive real-robot experiments across diverse articulated-object categories and interaction configurations, demonstrating the effectiveness of GSAM in terms of perception accuracy, overall success rate, and efficiency.
\end{itemize}
The remainder of this paper is organized as follows. Section 2 reviews related work, Section 3 presents the GSAM framework, Section 4 reports the experimental results, and Section 5 concludes the paper with limitations and future directions.
\section{Related Work}
\textbf{Policy Learning}. Due to the close link between robotic manipulation and human behavior, policy learning has been widely explored in articulated object manipulation. It involves imitation learning (IL)~\cite{chi2024umi,fan2023digital,wu2024gello,levine2018learning}, which learns from human demonstrations, and reinforcement learning (RL)~\cite{kalashnikov2018scalable}, which improves robustness via simulation. Some IL works~\cite{bahl2022human,ma2023vip,wang2023mimicplay} learn policies from in-the-wild human videos, but domain gaps hinder transfer to robotic settings. To address this, \cite{chi2023diffusion,wong2022error,seo2023deep} employ teleoperated robot interfaces, e.g., 3D-space mouse, smartphones, VR/AR, to collect task-specific data, though they are often costly or unintuitive due to latency issues. ALOHA~\cite{fu2024mobile} and GELLO~\cite{wu2023gello} balance cost and intuitiveness but still rely on special hardware. Recent methods~\cite{zhao2023learning,chi2023diffusion} use handheld devices mimicking grippers for data collection. However, IL approaches still require large expert datasets, even for simple tasks. RL~\cite{kalashnikov2018scalable} has been introduced to improve policy robustness, but is mostly limited to simulation. In contrast, our method follows a modular paradigm, where perception, interaction, and manipulation modules operate independently yet cooperatively, improving generalization to diverse objects.

\noindent \textbf{Vision-Motion Planning.} 
Vision-motion planning methods are popular for their flexibility in integrating object perception and interaction. They typically begin with visual estimation of kinematic parameters, followed by trajectory planning for the object interaction. Perception accuracy heavily depends on input types~\cite{yu2024gamma,ehsani2021manipulathor,xia2024kinematic}, commonly including point clouds~\cite{liu2023paris,Wang2025articubot} and RGB-D images~\cite{qian2023understanding,jiang2022opd,gupta2025openingarticulatedobjectsreal}. High-quality point clouds require multi-view fusion using sensors from different angles, whereas RGB-D images can be captured via wrist-mounted cameras, making them more convenient for real-world collection. Existing models struggle with kinematic parameters identification. Some~\cite{sun2023opdmulti,jiang2022opd} use pixel-level regression, while others~\cite{qian2023understanding} apply polar coordinates. However, all of them often ignore commonsense cues, leading to biased estimations. Moreover, traditional motion planners~\cite{karaman2011sampling,kavraki1996probabilistic} ensure reachability but overlook fine-grained interactions, risking collisions in narrow spaces. To address these issues, our method employs RGB-D inputs~\cite{qian2023understanding}, refines kinematic parameters via VLM's commonsense-reasoning, and incorporates interaction constraints through LLM to enhance safety during the whole manipulation procedure.

\noindent \textbf{LLM/VLM Guidance.} Recent methods~\cite{kim2024survey,su2025ova-fields,wang2025ske-layout} have leveraged LLMs and VLMs to incorporate world knowledge into robotic manipulation. These models can directly generate keypoints and trajectory constraints for the end-effector~\cite{huang2024rekep,yao2025sim2real}, enabling adaptability to complex environments. However, most works are limited to simple pick-and-place tasks. For articulated objects with rigid constraints, these methods often underperform. For instance,~\cite{xia2024kinematic} encoded kinematic parameters into textual formats and used LLMs for trajectory planning, but this paradigm mainly excelled in simulation. In real-world scenarios, without incorporating object and environment knowledge, the limited spatial reasoning of LLM/VLM can lead to unsafe and unconstrained interactions, causing end-effector collisions. To mitigate this, our approach constructs a comprehensive knowledge base that integrates articulated object, interaction pose, and obstacle avoidance knowledge. This knowledge base is provided to LLMs as input to improve spatial reasoning and generate trajectories meeting the safety constraints.

\section{Method}
In this study, we primarily consider the task of a mobile manipulator opening 1-DoF articulated objects with diverse hinge type, size and surface texture, which is particularly prevalent in practical scenarios. We assume the object has a graspable handle to be opened from a fully closed state. Based on the above task statement and assumption, we aim to develop a robotic framework that takes the observations of the articulated object \(o\) as input and outputs a set of 3D trajectory waypoints representing the manipulation process.
\begin{figure*}[t]
  \centering
  \includegraphics[width=0.9\linewidth]{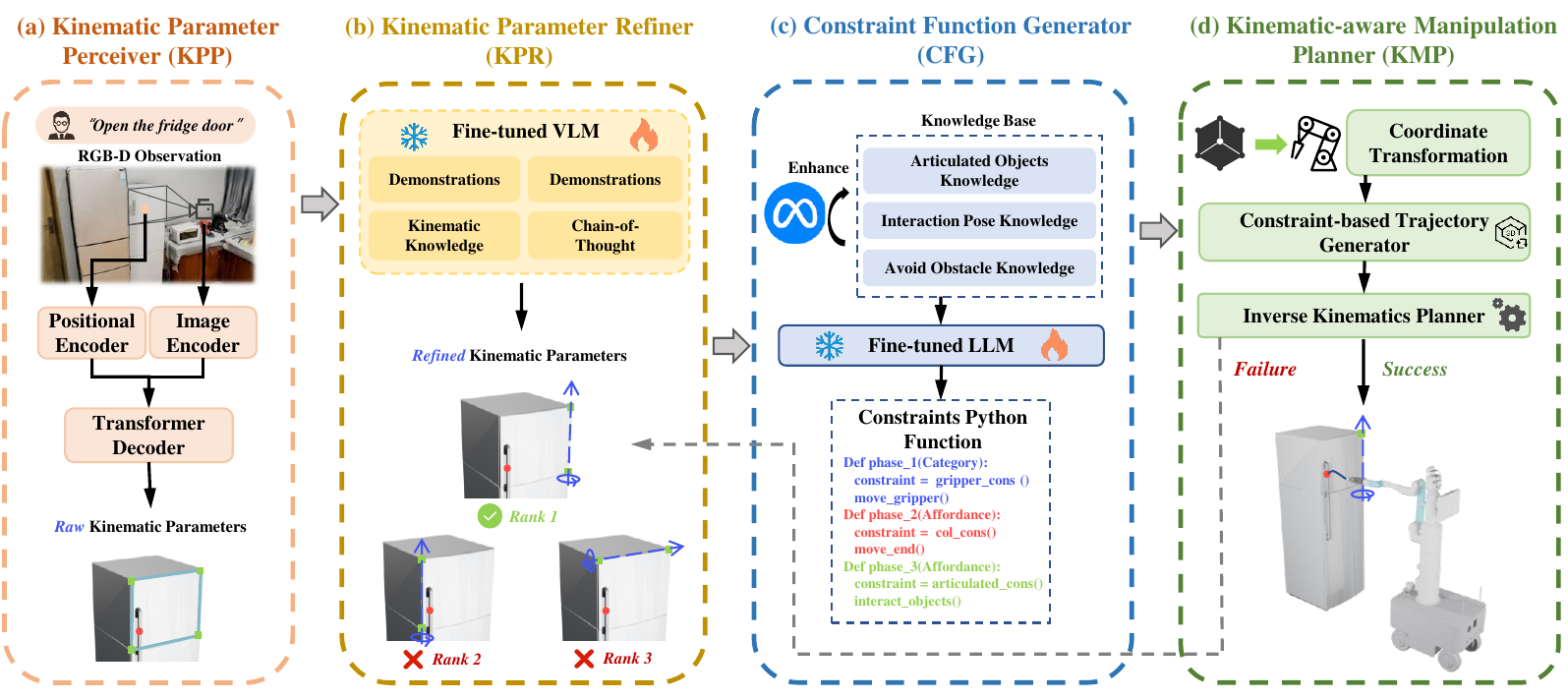} 
  \vspace{-6pt}
  \caption{GSAM overview.  KPP extracts visual features and decodes them into raw kinematic parameters. KPR refines them via VLM-based commonsense reasoning. The refined parameters and the constructed knowledge base are passed to CFG's LLM to generate Python constraint functions. KMP then converts these into end-effector coordinates and 3D waypoints to control mobile manipulators.}
  \label{fig:framework}
  \vspace{-12pt}
\end{figure*}

An overview of the GSAM framework is illustrated in Fig. \ref{fig:framework}. GSAM starts with the vision-based kinematic parameter perceiver (KPP) module. Then, it comprises two modules: the VLM-based kinematic parameters refiner (KPR) for credible {\em perception}, and the constraint function generator (CFG) for safe {\em interaction}. Finally, GSAM ends with the kinematic-aware manipulation planner (KMP) module for precise {\em execution}.

\subsection{Kinematic Parameter Perceiver}\label{sec:K2P}
As shown in Fig.~\ref{fig:framework}(a), the observation \(o\) consists of an RGB-D image and  keypoints within the image corresponding to the articulated object. Using SAM~\cite{kirillov2023segment} as our backbone, the KPP module estimates the raw kinematic parameters $K$ via two key steps, i.e., feature encoding and decoding. 
% Technical details are provided in Appendix~2.

\noindent \textbf{Encoder}. We employ the MAE~\cite{he2022masked} pre-trained Vision Transformer to encode the RGB-D image into a memory representation $m$. Then we use the positional encoder from~\cite{tancik2020fourier}, which combines Fourier mappings and MLPs to convert sparse prompts into high-dimensional embeddings.

\noindent \textbf{Decoder}. Following the architecture of 3DOI~\cite{qian2023understanding}, our Transformer decoder \(D\) takes memory \(m\), and a point query \(k_p\), and outputs a point-pooled feature \(h\):
\begin{equation}\label{eq:h1}
h = D(m; k_p).
\end{equation}
This feature \(h\) is then passed through {\em affordance, category, and mask} identification heads, which yield the interaction position, hinge type, and object size, respectively. These {identifications} are combined to form the raw kinematic parameters $K$.

\subsection{Kinematic Parameters Refiner}
\label{sec:KPR}
Although the KPP module can extract raw kinematic parameters from RGB-D observations, the initial predictions still suffer from keypoint drift, category ambiguity, or commonsense-inconsistent interaction locations. This is mainly caused by the structural diversity of articulated objects and the presence of occlusion, blurred boundaries, viewpoint bias in real scenes. 
In particular, for rotational objects, valid keypoints should lie close to the hinge axis, whereas for translational objects, the keypoints should jointly define a line or surface consistent with the sliding direction. Such structural relations are typically not directly inferable from isolated visual cues alone; instead, they necessitate multi-stage reasoning that integrates object categorization, handle-type identification, and interaction semantics.
% \begin{figure}[htbp]
%   \centering
%   \includegraphics[width=1\textwidth]{fig/Fig3.pdf} 
%   \vspace{-6pt}
%   \caption{Overview of KPR. {Top left}: Example demonstrations. {Top right}: Raw kinematic parameters and task instruction. {Center}: VLM prompt with demonstrations and chain-of-thought reasoning. {Bottom}: Refined outputs ranked by the confidence.}
%   \vspace{-12pt}
%   \label{fig:KPR_Prompt}
% \end{figure}
To address this issue, we introduce the Kinematic Parameter Refiner (KPR) after KPP to perform semantic enhancement and structural consistency correction over the raw keypoint set \(P\) and the corresponding kinematic parameter estimation \(K\), producing refined kinematic parameters \(K_{\text{refined}}\). As shown in Fig.~\ref{fig:framework}(b), KPR adopts a parameter-efficient fine-tuned open-source vision-language model to balance domain adaptability, controllability, and deployment cost. Specifically, we employ a low-rank adaptation (LoRA) fine-tuned LLaVA-7B as the refinement backbone and design task-specific correction prompts to guide keypoint filtering, completion, and parameter refinement.

Formally, the refinement process is defined as:
\begin{equation}
K_{\text{refined}} = \mathcal{M}_{\text{ref}}\big(o,\,K,\,\pi_{\text{corr}}\big).
\end{equation}
where \(o\) denotes the visual observation, \(K\) is the raw kinematic parameter set predicted by KPP, \(\pi_{\text{corr}}\) is the correction prompt set, and \(\mathcal{M}_{\text{ref}}\) denotes the fine-tuned refinement model.

The correction prompt set \(\pi_{\text{corr}}\) consists of four components:
\begin{equation}
\pi_{\text{corr}} = \{D_{\text{demo}},\,K,\,I_{\text{task}},\,P_{\text{cot}}\}.
\end{equation}
Here, \(D_{\text{demo}}\) denotes demonstrations, \(K\) denotes the raw kinematic parameters, \(I_{\text{task}}\) denotes the task instruction, and \(P_{\text{cot}}\) denotes the COT textual prompt.

The core motivation of this prompt design is that directly applying a VLM to filter \(P\) is unreliable, since effective keypoint refinement requires explicit decomposition of the reasoning process. For instance, valid keypoints on rotational objects are expected to co-align with the hinge-axis structure, while those on translational objects must exhibit geometric consistency with the sliding surface or motion direction. Critically, neither relation can be reliably inferred from isolated local visual cues alone. Inspired by~\cite{wei2022chain}, we therefore introduce a COT prompt that explicitly guides the model to first identify the object category, then infer the handle type around the affordance region, and finally refine the candidate keypoints by combining visual evidence with structural priors and commonsense knowledge. Through this stepwise reasoning process, the model moves beyond direct pattern matching and establishes a more reliable inference chain among object category, interaction region, and keypoint structure, thereby improving both keypoint selection quality and kinematic parameter accuracy.

To avoid the computational overhead of full-parameter adaptation, we adopt LoRA for parameter-efficient fine-tuning. During training, we freeze the visual encoder and most language model parameters, and inject low-rank adaptation matrices into the attention projection layers and selected feed-forward layers. Supervised instruction tuning is then used to learn the mapping from raw parameters to refined outputs. The training objective is formulated as the standard autoregressive cross-entropy loss:
\begin{equation}
L_{\text{sft}} = - \sum_{t=1}^{T} \log p\big(y_t \mid y_{<t},\, o,\, K,\, \pi_{\text{corr}}\big),
\end{equation}
where \(y_t\) denotes the target token at time step \(t\), and \(T\) is the output sequence length.

Given the diversity of articulated objects, KPR does not output only a single refined solution. Instead, it produces multiple candidate keypoint sets and their corresponding candidate refined kinematic parameters, which are further ranked according to structural consistency, interaction plausibility, and commonsense compatibility. The final output is thus a ranked candidate set:
\begin{equation}
K_{\text{refined}} = \left\{K_{\text{refined}}^{(1)},\,K_{\text{refined}}^{(2)},\,\dots,\,K_{\text{refined}}^{(M)}\right\},
\end{equation}
where \(K_{\text{refined}}^{(1)}\) denotes the highest-confidence candidate, and \(M\) is the number of retained candidates.

Overall, the introduction of KPR endows GSAM with structural-prior-aware and semantically grounded refinement capability on top of the raw visual estimation. By parameter-efficiently fine-tuning an open-source vision-language model for articulated object manipulation, the system can reliably map raw kinematic predictions to refined parameter candidates without incurring substantial deployment cost, thus providing more accurate and consistent kinematic inputs for the subsequent constraint generation module.
% The KPR module uses VLM with rich structural priors to update the keypoint set \(P\), producing a refined set and more accurate kinematic parameters \(K_{\text{refined}}\). This can be seen in Fig.~\ref{fig:framework}(b). Specifically, we employ GPT-4o as the VLM and design task-specific refinement prompts to guide keypoint selection and enhancement.
% The process is defined as:
% \begin{equation}
% K_{\text{refined}} = \text{VLM}\big(K,\,\text{refined\_prompts}\big).
% \end{equation}
% As exhibited in Fig.~\ref{fig:KPR_Prompt}, \(refined\_prompts\) consists of four components: demonstrations, raw kinematic parameters, task instructions, and chain-of-thought textual prompt. Directly using a VLM is hard to filter \( P \) well, as effective filtering requires multi-step spatial reasoning. For instance, keypoints for rotational objects correspond to points along the rotation axis, whereas those for translational objects define a surface plane—both are not directly observable. Without contextual decomposition, the VLM often fails to ground its identifications. Inspired by \cite{wei2022chain}, we propose a chain-of-thought prompt that first guides the VLM to consider the object category and classify the handle type at the affordance location. This information enables the VLM to use structural priors to filter candidate points and select appropriate keypoints. Given the diversity of articulated objects, multiple candidate keypoint sets are generated as potential solutions, each corresponding to a distinct set of candidate kinematic parameters \(K_{\text{refined}}\).
\subsection{Constraint Function Generator}\label{sec:CFG}
CFG retrieves information from the pre-constructed knowledge base $B$ and generates trajectory constraint functions $C$ tailored for articulated object manipulation. Furthermore, $B$ is progressively enriched through each execution, enabling continual refinement of the reasoning and planning capabilities of the system. The CFG is organized around two core components: the construction of $B$ and query-based constraint generation. %as shown in Fig.~\ref{fig:CFG}.

% \begin{figure}[htbp]
%   \centering
%   \includegraphics[width=1\textwidth]{fig/Fig4.pdf} 
%   \vspace{-6pt}
%   \caption{Overview of CFG. Knowledge base \(B\) include articulated objects, interaction pose, and obstacle avoidance knowledge. LLM receives the refined kinematic parameters \(K_{\text{refined}}\) along with chain-of-thought prompt, queries \(B\) to retrieve relevant information, and synthesizes executable constraint function \(C\). After successful manipulation, the newly generated knowledge is used to augment \(B\).}
%   \vspace{-12pt}
%   \label{fig:CFG}
% \end{figure}

\noindent \textbf{Knowledge Base Construction}. 
We partition knowledge base $B$ into three subsets $B_{\mathrm{int}}$, $B_{\mathrm{obs}}$, $B_{\mathrm{art}}$, expressed as:
\begin{equation}\label{eq:f_Kownledge}
B = \{ B_\text{int},\,B_\text{obs},\,B_\text{art} \}.
\end{equation}
\begin{itemize}
  \item \textit{Interaction Pose Knowledge} ($B_{\text{int}}$). We define each entry as an ordered pair $(c, g)$, where $c$ is the textual handle description, and $g$ is the corresponding end-effector pose, represented as the triplet $(\phi, \theta, \psi)$. Here, $\phi$, $\theta$, and $\psi$ denote the roll, pitch, and yaw angles of the end-effector, respectively. Each pair is encoded as a function and stored in \(B_{\mathrm{int}}\).

  \item \textit{Obstacle Avoidance Knowledge} ($B_{\text{obs}}$). We apply the sampling-based motion planning algorithm \(\mathrm{RRT}^*\)~\cite{karaman2011sampling} to generate collision-free trajectories toward the interaction point. Rather than storing explicit coordinates, we extract coordinate transformation patterns and encode them as functions in \(B_{\mathrm{obs}}\).

  \item \textit{Articulated Object Knowledge} ($B_{\text{art}}$). After reaching the interaction point and closing the gripper, we use ground-truth articulation parameters to guide the opening motion and record the resulting trajectory. The optimal trajectory is computed as:
    \begin{equation}
      \tau^*(q)
      = T_{\mathrm{obj}}(q)\,
        T_{\mathrm{obj}}^{-1}(q_{\mathrm{init}})\,
        T^{\mathrm{init}}_{\mathrm{eef}},
    \end{equation}
  where \(\tau^*(q)\) denotes the desired end-effector pose trajectory parameterized by joint parameter \(q\), which is either a rotation angle (revolute joint) or linear displacement (prismatic joint). \(q_{\mathrm{init}}\) is the joint value at initial grasp. We then perform least-squares fitting to match predicted poses to true recorded poses:
\begin{equation}
  f_{\mathrm{open}}
  = \arg\min_{f}
    \sum_{j=1}^{M}
    \bigl\lVert\,f(\mathbf{p}_{j}) - \boldsymbol{\tau}_{j}\bigr\rVert^{2},
\end{equation}
\begin{equation}
  B_{\mathrm{art}} = \{\,f_{\mathrm{open}}\,\},
\end{equation}
  where \(M\) is the number of trajectory samples, \(\mathbf{p}_j\) is the predicted pose from the proper trajectory, and \(\boldsymbol{\tau}_j\) is the corresponding recorded pose. The fitted function \(f\) maps predicted poses to ground-truth poses.
\end{itemize}
$B_{\mathrm{int}}$, $B_{\mathrm{obs}}$, and $B_{\mathrm{art}}$ respectively correspond to the end-effector {\em interaction, approaching, and opening} processes of articulated objects. We construct them from simulation environments~\cite{Yang2024BestManAM,mo2019partnet}, covering 60 instances across 5 articulated object types.

\noindent \textbf{Constraint Generation}. CFG maps the refined kinematic parameters \(K_{\text{refined}}\) to a set of executable trajectory constraints \(C\) via a query-based generation mechanism. We implement CFG with a parameter-efficient fine-tuned Llama 3.1-8B-Instruct model, adapted by LoRA to generate constraint-function text conditioned on the task description, refined kinematic parameters, and retrieved knowledge entries. Formally,
\begin{equation}
C = \mathcal{M}_{\text{cfg}}\big(K_{\text{refined}},\,B,\,\pi_{c}\big),
\end{equation}
where \(\mathcal{M}_{\text{cfg}}\) denotes the fine-tuned instruction model, and \(\pi_{c}\) is the COT prompt.

The prompt explicitly decomposes the articulated manipulation task into multiple sub-stages such as interaction-pose adjustment, approach and opening, and guides the model to retrieve relevant knowledge from \(B_{\text{int}}\), \(B_{\text{obs}}\), and \(B_{\text{art}}\) for stage-wise constraint generation. This stepwise reasoning improves both interpretability and executability.

We fine-tune the model on a dedicated dataset composed of structured parameters, constraint and knowledge entries. With most backbone parameters frozen and LoRA injected into the attention projection layers, the model is optimized by supervised instruction tuning:
\begin{equation}
L_{\text{cfg}} = - \sum_{t=1}^{T} \log p\big(y_t \mid y_{<t},\, o,\, K_{\text{refined}},\, B,\, \pi_{c}\big),
\end{equation}
where \(y_t\) is the output token at step \(t\). The resulting model can adaptively retrieve relevant knowledge and synthesize constraint functions according to the object category, interaction type, and environmental structure.
% \noindent \textbf{Query-based Constraint Generation}. A COT prompt feeds the refined kinematic parameters \(K_{\text{refined}}\) into LLM, which synthesizes the $C$ representing end-effector trajectory constraints. After each successful interaction, new information is added to the knowledge base $B$, enabling continual refinement for future planning.

\subsection{Kinematic-aware Manipulation Planner}\label{sec:KMP}
KMP aims to generate 3D waypoints in the mobile manipulator's coordinate frame, verify that they satisfy robot arm's kinematic constraints, and ultimately actuate the arm to execute all interactions with articulated objects.\\
\noindent \textbf{Coordinate Transformation and Trajectory Generation}. The image coordinates captured by the wrist-mounted camera are transformed into the mobile manipulator's coordinate frame, and then passed into the constraint-based trajectory generator to produce trajectory waypoints. The constraint-based trajectory generator employs minimum-snap optimization under $C$, partitioning the trajectory into $N$ polynomial segments and subsequently determining the optimal coefficients for each segment through numerical solution.\\ % Detailed transformation equations are provided in the Appendix.
\noindent \textbf{Inverse Kinematics}. The 3D waypoints $\{p_i\}$ produced by this process are fed into an inverse kinematics planner to compute the corresponding joint configurations and verify their reachability. Formally, for each waypoint $p_i$, we solve:
\begin{equation}
\label{eq:ik_all}
\begin{aligned}
q_i &= \arg\min_{q\in\mathbb{R}^n} \bigl\|\mathrm{FK}(q) - p_i\bigr\|^2,\\
\text{s.t.}\quad
q_{\min} &\le q \le q_{\max},\quad
\mathrm{CollisionFree}(q).
\end{aligned}
\end{equation}
Here, $\mathrm{FK}(q)$ denotes the forward-kinematics mapping, $q_{\min}, q_{\max}$ are joint limits, and $\mathrm{CollisionFree}(q)$ enforces no collisions. The solutions $\{q_i\}$ represent joint-space configurations for waypoints $\{p_i\}$ that satisfy the robot arm's joint-limit and kinematic constraints. In practice, this inverse kinematics transformation is implemented using the Open Motion Planning Library (OMPL).\\
\noindent \textbf{Rollback Strategy}. Considering that the design of some articulated objects in the real world may deviate from common assumptions, our predicted trajectories may not always be fully accurate. For example, a door's hinge axis is typically assumed to lie on the right or left, but it may actually be located at the top. To this end, we implement a rollback strategy: once an interaction failure occurs, we sequentially invoke the remaining kinematic knowledge in the KPR module to generate updated constraints, and retry the interaction with the current articulated object. This rollback will loop continuously until a successful manipulation is achieved or all kinematic parameter candidates have been retried.

\section{Experiments}
\subsection{Experimental Setup}
\noindent \textbf{Environment.} 
We conduct 50 real-robot manipulation trials in total. The benchmark contains five unseen articulated-object categories: right-hinge, left-hinge, bottom-hinge, prismatic, and textured-hinge objects. For each category, five object instances are selected, and each instance is tested from two base placements, resulting in $5$ categories $\times$ $5$ instances $\times$ $2$ placements $=$ $50$ trials. The same 50 trials are used for all compared methods.

The hardware platform consists of a Ranger Mini 3.0 mobile base, a custom metal body, and a Realman 63-F robotic arm. An Intel RealSense D435i camera is mounted on the wrist to capture RGB-D images. An NVIDIA RTX 3090 GPU is configured in the system to support computational requirements.

\noindent \textbf{Baselines.}
We compare GSAM with three representative categories. (1) LLM/VLM guidance: Kinematic LLM, and A3VLM. (2) Policy learning: Behavior Cloning. (3) Vision-motion planning: 3DOI+Motion Planning.
\begin{itemize}
    \item \emph{Kinematic LLM}~\cite{xia2024kinematic}: Uses LLMs to generate 3D waypoints directly.

    \item \emph{A3VLM}~\cite{huang2024a3vlm}: Uses VLM to predict kinematic knowledge then generate 3D waypoints.
    
    \item \emph{Behavior Cloning}~\cite{chi2023diffusion,chi2024umi}: Works as an end-to-end imitation learning approach.
    
    \item \textcolor{black}{\emph{3DOI+Motion Planning}}: Extracts kinematic parameters with the vision-based 3DOI method~\cite{qian2023understanding}, and performs interaction through motion planning using RRT*~\cite{chen2024rbirrts}.
    
    \item \textcolor{black}{\emph{OPD+Motion Planning}}: Obtains kinematic parameters via the OPD model~\cite{jiang2022opd}, followed by interaction planning using RRT*~\cite{chen2024rbirrts}.
    
\end{itemize}
\vspace{-8pt}
\begin{table*}[htbp]
    \caption{{\color{black}Comparison of GSAM against baseline methods and ablation study in terms of PSR, ISR, and OSR (\%). Results are reported for each hinge type and averaged across all 50 tasks in campus environments.}}
    \centering
    \normalsize
    \setlength{\tabcolsep}{1.5pt}
    \renewcommand{\arraystretch}{1.3}
    \resizebox{\linewidth}{!}{%
    \begin{tabular}{
        l
        *{15}{c}
        |
        c c c
    } \hline
        {Methods} & 
        \multicolumn{3}{c}{Right Hinge} & 
        \multicolumn{3}{c}{Prismatic Hinge} & 
        \multicolumn{3}{c}{Bottom Hinge} & 
        \multicolumn{3}{c}{Left Hinge}  & 
        \multicolumn{3}{c}{Textured Hinge} & 
        \multicolumn{3}{c}{Mean $\pm$ Std} 
         \\ \hline
        & PSR & ISR & OSR
        & PSR & ISR & OSR
        & PSR & ISR & OSR
        & PSR & ISR & OSR
        & PSR & ISR & OSR
        & PSR & ISR & OSR\\ \hline
        \multicolumn{19}{c}{\textbf{GSAM vs Baselines}} \\
        \emph{Kinematic LLM} & 
        90.0 & 11.1 & 10.0 & 
        100.0 & 60.0 & 60.0 & 
        100.0 & 20.0 & 20.0 &
        90.0 & 44.4 & 40.0 &
        80.0 & 25.0 & 20.0 &
        \cellcolor{gray!20}92.0 $\pm$ 7.5 & \cellcolor{gray!20}31.2 $\pm$ 18.3 & \cellcolor{gray!20}30.0 $\pm$ 16.3 \\
        \emph{A3VLM} & 
        50.0 & 60.0 & 30.0 &
        90.0 & 77.8 & 70.0 &
        60.0 & 66.7 & 40.0 &
        50.0 & 20.0 & 10.0 &
        60.0 & 66.7 & 40.0 &
        \cellcolor{gray!20}62.0 $\pm$ 14.7 & 
        \cellcolor{gray!20}61.3 $\pm$ 19.9 & 
        \cellcolor{gray!20}38.0 $\pm$ 19.4 \\
        \emph{Behavior Cloning} & 
        / & / & 20.0 &
        / & / & 30.0 &
        / & / & 20.0 &
        / & / & 30.0 &
        / & / & 30.0 &
        \cellcolor{gray!20}/ & 
        \cellcolor{gray!20}/ & 
        \cellcolor{gray!20}26.0 $\pm$ \textbf{4.5} \\
        \emph{3DOI+Motion Planning} & 
        50.0 & 80.0 & 40.0 &
        100.0 & 70.0 & 70.0 &
        70.0 & 71.4 & 50.0 &
        60.0 & 66.7 & 40.0 &
        70.0 & 85.7 & 60.0 &
        \cellcolor{gray!20}70.0 $\pm$ 16.7 & 
        \cellcolor{gray!20}74.8 $\pm$ 7.0 & 
        \cellcolor{gray!20}52.0 $\pm$ 10.6 \\
        \emph{OPD+Motion Planning} & 
        40.0 & 75.0 & 30.0 &
        70.0 & 71.4 & 50.0 &
        / & / & / &
        50.0 & 60.0 & 30.0 & 
        40.0 & 75.0 & 30.0 &
        \cellcolor{gray!20}50.0 $\pm$ 12.2 & 
        \cellcolor{gray!20}70.0 $\pm$ 6.2 & 
        \cellcolor{gray!20}35.0 $\pm$ 14.8 \\
        \textbf{\emph{GSAM} (Ours)} & 
        90.0 & 88.9 & \textbf{80.0} &
        100.0 & 100.0 & \textbf{100.0} &
        100.0 & 90.0 & \textbf{90.0} &
        90.0 & 100.0 & \textbf{90.0} &
        80.0 & 100.0 & \textbf{80.0} &
        \cellcolor{gray!20}\textbf{92.0 $\pm$ 7.5} & 
        \cellcolor{gray!20}\textbf{95.7 $\pm$ 5.2} & 
        \cellcolor{gray!20}\textbf{88.0} $\pm$ 7.5 \\
        \hline
        \multicolumn{19}{c}{\textbf{Ablation Study}} \\
        \emph{GSAM w/o KPR} & 
        50.0 & 100.0 & 50.0 &
        100.0 & 100.0 & 100.0 &
        60.0 & 83.3 & 50.0 &
        70.0 & 71.4 & 50.0 &
        60.0 & 100.0 & 60.0 &
        \cellcolor{gray!20}68.0 $\pm$ 17.2 & 
        \cellcolor{gray!20}91.0 $\pm$ 11.7 & 
        \cellcolor{gray!20}62.0 $\pm$ 19.4 \\ 
        \emph{GSAM w/o CFG} & 
        90.0 & 33.3 & 30.0 &
        100.0 & 60.0 & 60.0 &
        100.0 & 40.0 & 40.0 &
        90.0 & 60.0 & 50.0 &
        80.0 & 37.5 & 30.0 &
        \cellcolor{gray!20}92.0 $\pm$ 7.5 & 
        \cellcolor{gray!20}44.2 $\pm$ 9.6 & 
        \cellcolor{gray!20}42.0 $\pm$ 11.7 \\
        \emph{GSAM w/o COT} & 
        60.0 & 66.7 & 40.0 &
        100.0 & 80.0 & 80.0 &
        70.0 & 57.1 & 40.0 &
        70.0 & 71.4 & 50.0 &
        80.0 & 50.0 & 40.0 &
        \cellcolor{gray!20}76.0 $\pm$ 13.6 & 
        \cellcolor{gray!20}65.0 $\pm$ 10.5 & 
        \cellcolor{gray!20}50.0 $\pm$ 15.5 \\
    \hline
    \end{tabular}}
    \label{tab:main}
    \vspace{-15pt}
\end{table*}
\vspace{-8pt}

\noindent \textbf{Evaluation.}
We adopt Perception Success Rate (\(\rm{PSR}\)), Interaction Success Rate (\(\rm{ISR}\)), and Overall Success Rate (\(\rm{OSR}\)) as the evaluation metrics to study the generalization and safety of all articulated object manipulation methods. Specifically, $\rm{PSR}$ is defined as $\rm{P_{\mathrm{success}}}/{N_{\mathrm{total}}}$, where \({\rm{P}}_{\mathrm{success}}\) is the number of successful perception cases,and \(\rm{N}_{\mathrm{total}}\) is the total number of tasks.
Interaction Success Rate (\(\rm{ISR}\)) is defined as $\rm{I_{\mathrm{success}}}/{P_{\mathrm{success}}}$, where \({\rm{I}}_{\mathrm{success}}\) represents successful interactions without collision. For revolute hinges, a success requires an opening angle of at least \(60^\circ\); for prismatic hinges, it needs at least \(20\) cm of translation. Given sequential nature of perception and interaction, Overall Success Rate $\rm{OSR}$ is defined as $\rm{PSR \times ISR}$. Higher $\rm{PSR}$, $\rm{ISR}$, and $\rm{OSR}$ indicate better performance.
Besides, to assess the efficiency and practical values of all methods, we report the Task Execution Time (\(\rm{Time}\)). 

\subsection{Numerical Results}
Numerical results are presented in Table \ref{tab:main}, where ``/'' means that steps in some baselines cannot be executed independently. Since Behavior Cloning is trained in an end-to-end manner, it can only be measured by the average \(\rm{OSR}\). GSAM and Kinematic LLM achieve the same \(\rm{PSR}\), attributable to Kinematic LLM's absence of visual perception and its dependence on kinematic parameters estimated by our perception module.

\noindent \textbf{Generalization Results.}
% On 5 different categories of articulated objects, the $\rm{OSR}$ for GSAM, Kinematic LLM, A3VLM, Behavior Cloning, 3DOI+Motion Planning, and OPD+Motion Planning fall within {\color{black}[80.0\%,100.0\%], [10.0\%,60.0\%],  [10.0\%,70.0\%], [20.0\%,30.0\%], [40.0\%,\allowbreak 70.0\%], and [0\%,50.0\%], with standard deviations of 7.5\%, 16.3\%, 19.4\%, 4.5\%, 10.6\%, and 14.8\%, respectively.} 
Across the five categories, GSAM achieves the highest OSR (88.0\%) while maintaining a relatively small standard deviation (7.5\%).
(1) GSAM maintains a relatively small standard deviation with highest $\rm{OSR}$ across all categories. This is due to the commonsense reasoning in KPR. Even under irregular spatial configurations, it helps uncover structural cues from images and couple mask and axis identification, enabling accurate kinematic parameter estimation. (2) Most baselines, except Behavior Cloning, show large $\rm{OSR}$ fluctuations across categories. They perform best on prismatic hinge tasks, but drop sharply on right-, bottom-, left-, and textured-hinge tasks. This is partly because prismatic tasks usually present more regular handle layouts, making segmentation and kinematic estimation easier. In contrast, irregular hinge layouts increase spatial errors in 3DOI+Motion Planning and OPD+Motion Planning, lead to inaccurate waypoint generation in Kinematic LLM, and reduce the robustness of A3VLM due to its reliance on scene-extrema depth denormalization and single-shot regression. (3) Although Behavior Cloning has low variance, its $\rm{OSR}$ remains consistently low because imitation learning generalizes poorly to unseen objects.

\noindent \textbf{{\color{black}Safety Results.}} 
{\color{black} On 50 tasks with randomly initialized interaction configurations, the average manipulation success rate of GSAM is 88.0\%, while that of the optimal baseline 3DOI+Motion Planning is only 52.0\%. (1) GSAM outperforms all baselines in $\rm{OSR}$ across all interaction configurations. Its performance stems from a structured knowledge base integrating articulated object, interaction pose, and obstacle avoidance information. CFG further synthesizes constraint-aware trajectories, effectively mitigating collisions. (2) 3DOI+Motion Planning, and OPD+Motion Planning perform better than the remaining baselines, but their collision avoidance ability needs to be further improved. They often overlook the geometric structure of the handle in the interaction stage, increasing collision risk. In addition, when addressing bottom hinge tasks, OPD+Motion Planning fails, since OPD can not estimate the objects' kinematic parameters. (3) Kinematic LLM, \color{black}{A3VLM} and Behavior Cloning perform poorly. Without geometric or physical awareness, 3D waypoints generated by Kinematic LLM and A3VLM lead to collisions due to hallucinations. The distribution consistence requirement between training data and test objects makes Behavior Cloning difficult to output feasible trajectory under varying interaction configurations.}
\begin{table*}[htbp]
    \caption{{\color{black}Comparison of GSAM against baseline methods and ablation study in terms of task execution time ($\rm{s}$). Results are reported for each hinge type and averaged across all 50 tasks in campus environments.}}
    \centering
    \footnotesize
    \setlength{\tabcolsep}{2pt}
    \renewcommand{\arraystretch}{1.05}
    \resizebox{\textwidth}{!}{%
    \begin{tabular}{
        l
        *{5}{c}
        |
        c
    } \hline
        {Methods} & 
        {Right Hinge} & 
        {Prismatic Hinge} & 
        {Bottom Hinge} & 
        {Left Hinge} & 
        {Textured Hinge} & 
        {Mean $\pm$ Std} 
         \\ \hline
        \multicolumn{7}{c}{\textbf{GSAM vs Baselines}} \\
        \emph{Kinematic LLM} & 
        37.3 & 47.4 & 57.9 & 56.6 & 62.7 &
        \cellcolor{gray!20}52.4 $\pm$ 10.1\\
        \emph{A3VLM} & 
        37.1 & 40.9 & 51.4 & 43.8 & 50.8 &
        \cellcolor{gray!20}44.8 $\pm$ 6.2\\
        \emph{Behavior Cloning} & 
        19.4 & 22.6 & 32.2 & 24.1 & 20.1 &
        \cellcolor{gray!20}\textbf{23.7} $\pm$ 5.1\\
        \emph{3DOI+Motion Planning} & 
        33.4 & 44.0 & 42.6 & 40.6 & 42.4 &
        \cellcolor{gray!20}40.6 $\pm$ \textbf{4.2}\\
        \emph{OPD+Motion Planning} & 
        39.3 & 47.8 & / & 48.0 & 50.5 &
        \cellcolor{gray!20}46.4 $\pm$ 4.9\\
        \textbf{\emph{GSAM} (Ours)} & 
        38.4 & 46.0 & 49.2 & 39.4 & 37.5 &
        \cellcolor{gray!20}42.1 $\pm$ 5.2\\
        \hline
        \multicolumn{7}{c}{\textbf{Ablation Study}} \\
        \emph{GSAM w/o KPR} & 
        35.4 & 42.2 & 45.0 & 36.3 & 34.6 &
        \cellcolor{gray!20}38.7 $\pm$ 4.6\\ 
        \emph{GSAM w/o CFG} & 
        37.8 & 45.2 & 48.4 & 38.7 & 36.9 &
        \cellcolor{gray!20}{41.4 $\pm$ 5.1}\\
        \emph{GSAM w/o COT} & 
        40.2 & 48.4 & 51.9 & 41.3 & 39.2 &
        \cellcolor{gray!20}44.2 $\pm$ 5.6\\
    \hline
    \end{tabular}%
    }
    \label{tab:main_time}
    \vspace{-15pt}
\end{table*}
\noindent \textbf{{\color{black}Time  Results.}} 
{\color{black}On 50 tasks with randomly initialized interaction configurations, Behavior Cloning is the fastest method, achieving $23.7 \pm 5.1$s, since it only requires a single feed-forward inference without explicit online planning. The other model-based baselines are noticeably slower, with Kinematic LLM, A3VLM, 3DOI, and ODP requiring $52.4 \pm 10.1$s, $44.8 \pm 6.2$s, $40.6 \pm 4.2$s, and $46.4 \pm 4.9$s, respectively. GSAM attains a runtime of $42.1 \pm 5.2$s, which is substantially faster than  Kinematic LLM and ODP and comparable to 3DOI and A3VLM, while still delivering the strongest task success rates. These results show that, although GSAM cannot match the speed of end-to-end Behavior Cloning, it remains highly competitive in wall-clock efficiency and achieves a more favorable balance between efficiency and accuracy. Such a balance is crucial for leveraging the practical value of articulated objects manipulation.}

\subsection{Visual Results}
{\color{black}Perception, interaction, and manipulation are essential components of articulated task execution. In the perception evaluation, we compare KPP+KPR with three advanced baselines: OPD, A3VLM, and 3DOI. As reported in Table~\ref{tab:mask}, perception performance is assessed using two metrics Mask IoU and Axis EA-Score, which measure the accuracy of object mask prediction and kinematic axis estimation, respectively. The results show that GSAM achieves the best overall performance on both metrics, demonstrating its superior ability to localize articulated object regions and estimate key kinematic parameters. This advantage mainly stems from the perception refinement capability introduced by KPR, which enhances structural understanding through commonsense reasoning. Among the baseline methods, 3DOI generally performs better than OPD and A3VLM. However, due to the lack of commonsense-guided refinement, the axes estimated by 3DOI still tend to deviate from the optimal positions, as observed in the prismatic hinge task. In addition, 3DOI is prone to identifying incorrect axes in several cases, including the right, bottom, and left hinge tasks. These results demonstrate that GSAM provides more accurate and robust perception performance than the advanced baselines.}

\begin{table*}[htbp]
    \caption{Comparison of GSAM with baseline methods. Mask IoU and Axis EA-Score are reported for each hinge category and aggregated as mean performance over all 50 campus-scale tasks.}
    \centering
    \footnotesize
    \setlength{\tabcolsep}{2pt}
    \renewcommand{\arraystretch}{1.05}
    \resizebox{\textwidth}{!}{%
    \begin{tabular}{
        l
        *{10}{c}
        |
        c c
    } \hline
        {Methods} & 
        \multicolumn{2}{c}{Right Hinge} & 
        \multicolumn{2}{c}{Prismatic Hinge} & 
        \multicolumn{2}{c}{Bottom Hinge} & 
        \multicolumn{2}{c}{Left Hinge}  & 
        \multicolumn{2}{c}{Textured Hinge} & 
        \multicolumn{2}{c}{Mean} 
         \\ \hline
        & Mask & Axis 
        & Mask & Axis 
        & Mask & Axis
        & Mask & Axis
        & Mask & Axis 
        & Mask & Axis\\ \hline

        \emph{A3VLM} & 
        / & 38.5 &
        / & / &
        / & 44.8 &
        / & 49.0 &
        / & 42.5 &
        \cellcolor{gray!20} / & 
        \cellcolor{gray!20} 43.7 \\

        \emph{3DOI} & 
        72.0 & 71.8 &
        80.3 & / &
        76.8 & 85.5 &
        73.4 & 70.0 &
        75.1 & 78.6 &
        \cellcolor{gray!20} 75.5 & 
        \cellcolor{gray!20} 76.4 \\

        \emph{OPD} & 
        57.0 & 34.1 &
        65.8 & / & 
        / & / & 
        62.3 & 42.5 &
        60.5 & 36.1 &
        \cellcolor{gray!20} 61.4 & 
        \cellcolor{gray!20} 37.5 \\

        \textbf{\emph{GSAM} (Ours)} & 
        79.8 & 82.0 & 
        90.2 & / & 
        87.5 & 93.2 &
        82.1 & 89.4 &
        81.7 & 84.6 &
        \cellcolor{gray!20}\textbf{84.3} & 
        \cellcolor{gray!20}\textbf{87.3} \\
    \hline
    \end{tabular}%
    }
    \label{tab:mask}
    \vspace{-15pt}
\end{table*}

\begin{figure}[htbp]
  \centering
  \includegraphics[width=0.94\linewidth]{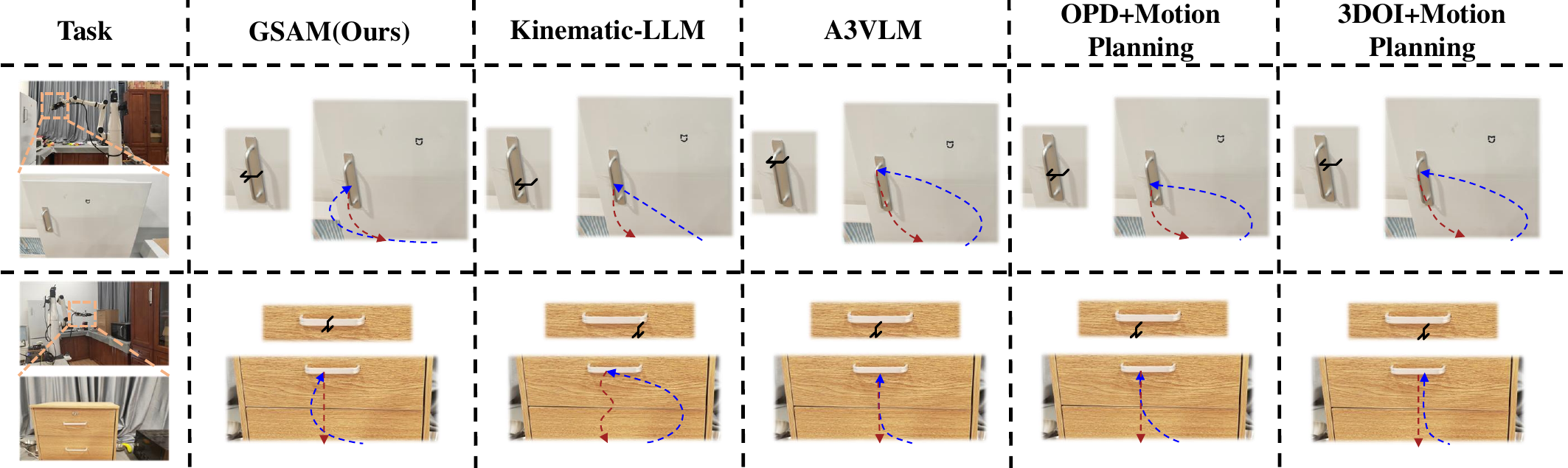} 
  \vspace{-6pt}
  \caption{Visual comparison of trajectory planning for interaction. Blue lines are approaching trajectories. {Red} lines denote opening trajectories. Local view illustrates interactions between end-effector and handle.}
  \vspace{-15pt}
  \label{fig:kinematic_GSAM}
\end{figure}

For interaction and manipulation, we compare CFG+KMP with Kinematic LLM, A3VLM, and Motion Planning baselines, where OPD+Motion Planning and 3DOI+Motion Planning use different vision-based kinematic parameters as inputs to the motion planner. As shown in Fig. \ref{fig:kinematic_GSAM}, GSAM generates intuitive and safe trajectories, whereas the baselines often suffer from interaction collisions. This advantage stems from CFG, which leverages a customized knowledge base to provide richer obstacle-avoidance constraints for trajectory planning.
\vspace{-8pt}

\subsection{Ablation Study}
To study the core modules KPR and CFG, as well as their COT questioning effectiveness, we conduct several degraded models. In GSAM w/o KPR, we remove the refiner KPR. In GSAM w/o CFG, we retain LLM but remove the assisted knowledge base. In GSAM w/o COT, we replace COT questioning with direct questioning in KPR and CFG. From Table~\ref{tab:main}, (1) in GSAM w/o KPR, removing KPR causes the average \(\rm{OSR}\) to drop from 88.0\% to 62.0\%. The absence of KPR prevents the VLM from performing unbiased kinematic parameter estimation, reducing $\rm{PSR}$ from 94.0\% to 68.0\% and leading to task failures. The 100\% ISR of GSAM w/o KPR is mainly because KPR is decoupled from the interaction stage; removing KPR only affects PSR, while ISR is determined by the downstream CFG and KMP modules. (2) In GSAM w/o CFG, removing knowledge base hinders the LLM’s ability to generate valid motion constraints, causing $\rm{ISR}$ to drop from 95.7\% to 44.2\% and $\rm{OSR}$ from 88.0\% to 42.0\%. (3) In GSAM w/o COT, removing COT prompting reduces $\rm{PSR}$ from 92.0\% to 76.0\% and $\rm{ISR}$ from 95.7\% to 65.0\%. COT externalizes intermediate reasoning, offering richer contextual cues and improving long-range dependency modeling. 

\subsection{Comparison to Point Cloud and Real2Sim2Real}
\begin{figure}[htbp]
  \centering
  \includegraphics[width=0.9\linewidth]{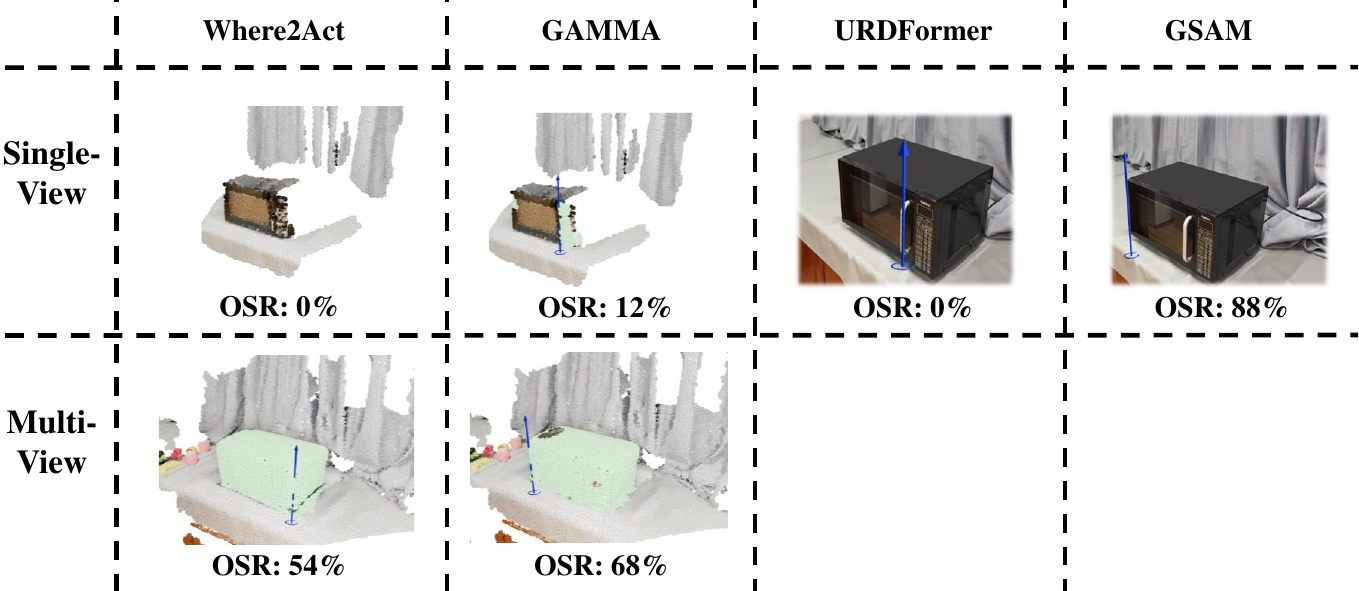} 
  \vspace{-6pt}
  \caption{Predicted segmentation and kinematic identification of a microwave.}
  \vspace{-15pt}
  \label{fig:vision1}
\end{figure}

We also investigate the performance gap between GSAM and state-of-the-art point cloud-based and Real2Sim2Real methods.  
% Specifically, we first use OpenMask3D~\cite{takmaz2023openmask3d} to obtain object-level segmentation, then apply Where2Act~\cite{mo2021where2act} and GAMMA~\cite{yu2024gamma} to implement kinematic parameter estimations and manipulation. Evaluations are conducted under both Single-View and Multi-View settings. In recent years, generative models introduce new paradigms for articulated object manipulation. We investigate the feasibility of this paradigm by employing URDFormer\cite{chen2024urdformer} to generate the required simulation assets.
As shown in Fig~\ref{fig:vision1}, GSAM consistently achieves higher $\rm{OSR}$ than Where2Act~\cite{mo2021where2act}, GAMMA~\cite{yu2024gamma}, and URDFormer\cite{chen2024urdformer} under both settings. (1) In the Single-View setting, point cloud-based methods often fail in kinematic estimation due to limited structural information and noisy point clouds, as shown in the upper part of Fig.~\ref{fig:vision1}. URDFormer performs particularly poorly, since its Real2Sim2Real pipeline produces distorted assets and misaligned reconstructions, leading to near-zero real-robot success. (2) Although Multi-View input improves geometry completeness and raises $\rm{OSR}$, both Single-View and Multi-View methods still underperform GSAM, as shown in the lower part of Fig.~\ref{fig:vision1}. This gap mainly arises because articulated objects are usually observed in closed states, where functional parts are hard to distinguish, making clustering and precise segmentation difficult. (3) GSAM bypasses these limitations by leveraging commonsense kinematic reasoning through VLM, enabling robust and reliable understanding and planning across varying scenarios.

\section{Conclusion}
In this study, we propose GSAM, a framework for generalizable and safe articulated object manipulation. GSAM introduces KPR to refine kinematic parameters via commonsense reasoning, improving generalization across diverse objects. CFG incorporates environmental constraints into trajectory generation, enhancing interaction safety. Extensive experiments show that, compared with the best baseline, GSAM improves generalization by 3.1\% across hinge tasks and reduces the collision rate by 36.0\% under random end-effector-handle configurations.

Despite its effectiveness, GSAM currently relies on monocular vision. Future work will integrate multimodal sensing, such as force feedback, to improve the reliability of closed-loop control. We will also extend GSAM to multi-DoF articulated objects.
\subsubsection{\ackname} This study was supported by the National Natural Science Foundation of China (Grant Nos. 62322601, 62572084 and 62506053), the Fundamental Research Funds for the Central Universities (Grant Nos. 2024IAIS-QN017 and 2025CDJZDGF001), and the Natural Science Foundation of Chongqing (Grant No. CSTB2024NSCQ-MSX0955).

\end{document}